\documentclass[journal]{IEEEtran}

\ifCLASSINFOpdf
\else
\fi

\ifCLASSOPTIONcompsoc
    \usepackage[caption=false, font=normalsize, labelfont=sf, textfont=sf]{subfig}
\else
	\usepackage[caption=false, font=footnotesize]{subfig}
\fi

\usepackage{cite}
\usepackage[version=4]{mhchem}
\usepackage{graphicx}
\usepackage{amsmath,amssymb,amsfonts}
\usepackage{color,soul}
\usepackage[table,xcdraw]{xcolor}
\usepackage{romannum}
\usepackage{graphicx}
\usepackage{commath}
\usepackage{textcomp}

\usepackage{booktabs}
\usepackage[utf8]{inputenc}
\usepackage{graphicx}
\usepackage{acronym}
\usepackage{authblk}
\usepackage{blindtext}
\usepackage{lettrine}
\usepackage{adjustbox}
\usepackage{amsmath}
\usepackage{cases}
\usepackage{xcolor}
\usepackage{cleveref}
\usepackage{pifont}
\usepackage{color, colortbl}
\definecolor{Gray}{gray}{0.9}
\ifCLASSOPTIONcompsoc
    \usepackage[caption=false, font=normalsize, labelfont=sf, textfont=sf]{subfig}
\else
\usepackage[caption=false, font=footnotesize]{subfig}
\fi
\usepackage{tabu}
\usepackage{tikz}

\usepackage{multirow}
\usepackage{threeparttable}

\usepackage{pdflscape}

\graphicspath{{figures/}}

\begin{document}
\title{Gradient-based Neuromorphic Learning on Dynamical RRAM Arrays}


\author{
Peng Zhou,~\IEEEmembership{Student Member,~IEEE,}
        Jason~K.~Eshraghian,~\IEEEmembership{Member,~IEEE,}
        Dong-Uk Choi,
        Wei~D.~Lu,~\IEEEmembership{Fellow,~IEEE},
        and Sung-Mo Kang,~\IEEEmembership{Life~Fellow,~IEEE}
        
\thanks{P. Zhou, D. Choi, and S. M. Kang are with the Department of Electrical and Computer Engineering, UC, Santa Cruz, CA, USA.}%
\thanks{J. K. Eshraghian and W. D. Lu are with the Department of Electrical Engineering and Computer Science, University of Michigan, Ann Arbor, MI, USA.}%
}



\maketitle

\begin{abstract}

We present \textit{MEMprop}, the adoption of gradient-based learning to train fully memristive spiking neural networks (MSNNs). Our approach harnesses intrinsic device dynamics to trigger naturally arising voltage spikes. These spikes emitted by memristive dynamics are analog in nature, and thus fully differentiable, which eliminates the need for surrogate gradient methods that are prevalent in the spiking neural network (SNN) literature.

Memristive neural networks typically either integrate memristors as synapses that map offline-trained networks, or otherwise rely on associative learning mechanisms to train networks of memristive neurons. We instead apply the backpropagation through time (BPTT) training algorithm directly on analog SPICE models of memristive neurons \textit{and} synapses. 
Our implementation is fully memristive, in that synaptic weights and spiking neurons are both integrated on resistive RAM (RRAM) arrays without the need for additional circuits to implement spiking dynamics, e.g., analog-to-digital converters (ADCs) or thresholded comparators. As a result, higher-order electrophysical effects are fully exploited to use the state-driven dynamics of memristive neurons at run time.
By moving towards non-approximate gradient-based learning, we obtain highly competitive accuracy amongst previously reported lightweight dense fully MSNNs on several benchmarks.

\end{abstract}

\begin{IEEEkeywords}
Neuromorphic computing, memristor, spiking neural network, supervised learning, backpropagation.
\end{IEEEkeywords}

\section{Introduction}

\begin{figure*}[!th]
  \begin{center}
  \includegraphics[width=7in]{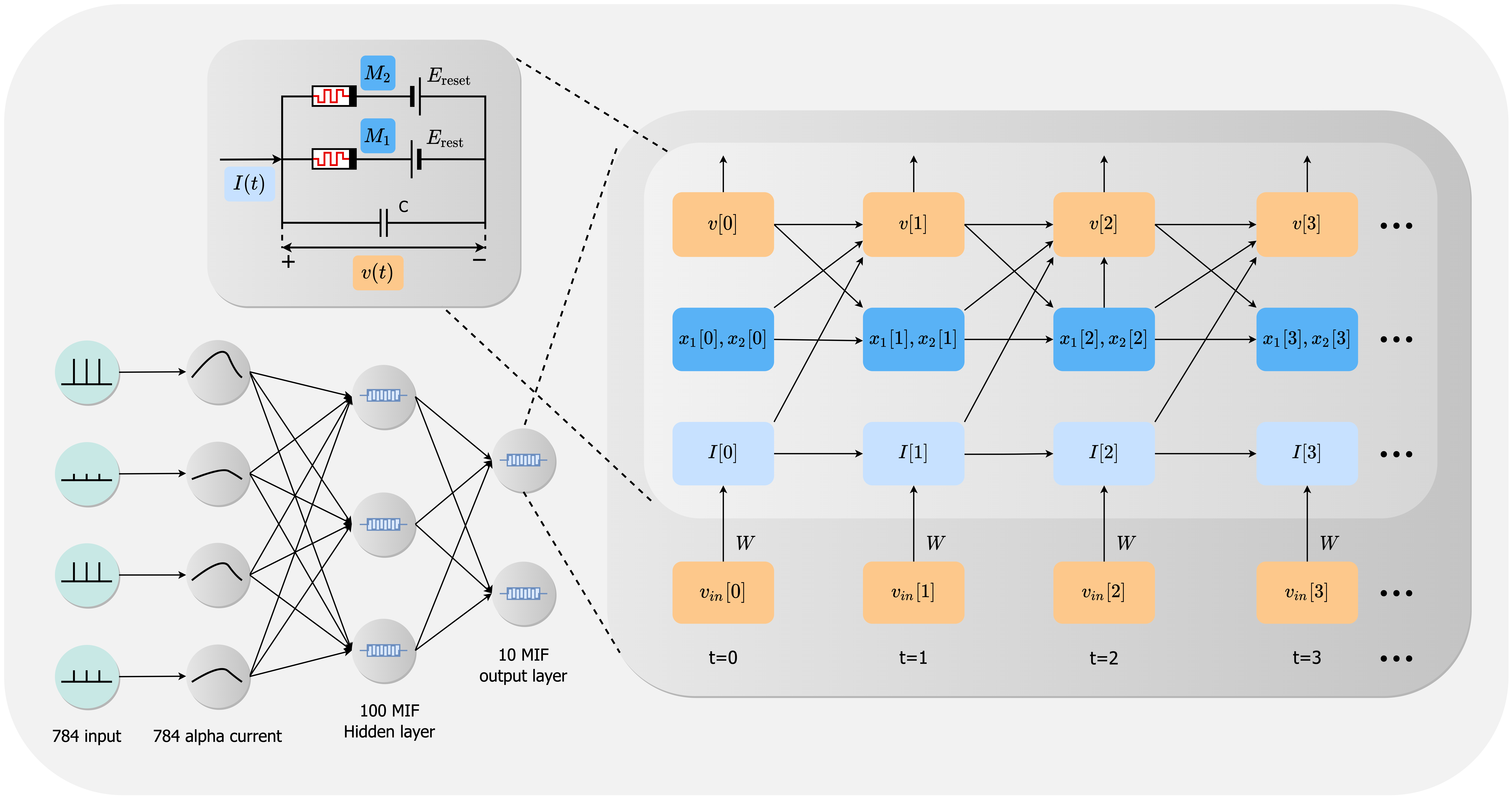}
  \caption{An overview of our MEMprop approach. The MSNN architecture and the resulting computational graph consisting of memristive dynamics.}\label{integrate}
  \end{center}
\end{figure*}

\IEEEPARstart{S}{piking} Neural Networks (SNNs) impose neuroscience-inspired constraints to modern deep learning algorithms, and have accordingly demonstrated significant improvements in runtime efficiency. By moving from full precision and fixed precision activations of artificial neuron models over to temporally-encoded data representations captured by spiking neurons, neuromorphic hardware has shown significant savings in energy consumption and latency~\cite{azghadi2020hardware, wang2020supervised, zhou2022fully, zhou2020towards, tavanaei2019deep, wunderlich2019demonstrating}.

The broad success of error backpropagation to train deep learning models has ushered in a plethora of related training algorithms adapted for SNNs, most of which are guided by surrogate gradient descent to overcome the non-differentiability of discrete spikes \cite{werbos1990backpropagation, neftci2019surrogate}. This proliferation of SNN usage is complemented by the development of modular deep learning programming packages that have optimized autodifferentiation for CUDA acceleration~\cite{snnTorch, sinabs, rockpool, hazan2018bindsnet, norse2021}.

In parallel to these advances in training SNNs, the past decade has seen huge strides in brain-inspired devices, circuits, and architectures that integrate neuronal dynamics to improve the hardware integration of SNNs and its constituent parts. Memristors and resistive RAM (RRAM) make up an immense part of such exploratory research in SNN implementation as they are a natural bridge between SNN algorithms and accelerators \cite{chua1971memristor, chua1976memristive}. They have been widely employed as both synapses and as spiking neurons.

At the ionic level, memristive synapses have been integrated into systems that naturally implement the spike-timing-dependent-plasticity (STDP) update rule using higher-order device dynamics \cite{jo2010nanoscale, serrano2013stdp, lin2020adaptive}. An alternative use of ion-driven dynamics is when implementing the memristor as a neuron, where nonlinear conductance evolution gives rise to abrupt switching that can be used to emit sudden voltage spikes. This approach is typically coupled with capacitive integration, and has been referred to as a `neuristor' \cite{pickett2013scalable, bao2019dual, del2020caloritronics, lim2015reliability}, and a `memristive integrate-and-fire' (MIF) neuron \cite{kang2021build, hao2020monolayer, zhou2022fully}. Similarly, membrane leakage in biological neurons can be implemented using resistive dissipation as in neuristors, or via volatile ionic drifting dynamics observed in single devices \cite{zhu2020memristor} and also in nanowire networks \cite{loeffler2021modularity, zhu2021mnist,hochstetter2021avalanches}.

Moving up to the architectural level, RRAM has been shown as a promising candidate for compute-in-memory (CIM) architectures due to their ability to parallelize matrix-vector multiplication independently of time complexity when integrated as large-scale, modular arrays \cite{cai2019fully, eshraghian20213, zhang2020brain, li2018analog, eshraghian2019analog}. Rather than neurons, memristive synapses map neural network weights to device conductances. In general, RRAM CIM architectures are intended to be trained offline with weights mapped on-chip for inference and deployment. As such, RRAM synapses should be stationary and only used for weight read-out. Higher-order dynamical behaviors of memristors are abstracted away, and treated as non-idealities.

An additional challenge with RRAM-based CIM is the cost of communicating analog current signals along lengthy bit-lines and conversion into the digital domain. These issues have spurred the use of binary activations in the form of spike-based CIM accelerators which have shown to alleviate the burdens of mixed-signal computation, by removing the need for large analog-to-digital (ADC) data converters \cite{eshraghian2022memristor}.

The majority of deep learning acceleration using memristors can be classified into one of the above cases: memristive neurons, memristive synapses that learn via associative learning, and CIM accelerators. A small set of designs have combined memristive neurons and memristive synapses together \cite{wang2018fully, tang2020fully}. Doing so is a commendable feat, as the natural switching dynamics of memristive systems are relied on to achieve data-driven tasks. The cost of allowing hardware behave naturally is that a designer can no longer rely on synchronous, clock-driven processing, and is subject to fault injections that are a result of nonlinear ionic dynamics. Letting the natural dynamics of memristive hardware to `teach itself' exacerbates the challenges of training MSNNs. This challenge has limited the demonstration of fully memristive SNNs to unsupervised learning tasks that have shown to solve simple, low-dimensional pattern recognition via local learning rules (typically STDP) and associative learning. These include the classification of several characters and numbers (such as a subset of the MNIST dataset). 

In this paper, we push beyond the classical limits of fully memristive SNNs, and demonstrate competitive performance on real-world datasets that go beyond static pattern recognition using a MSNN that includes both memristive synapses and memristive neurons. We achieve this by combining higher-order memristive dynamics with the gradient optimization process. While fault injections during weight update have been accounted for in the past, this is the first time nonlinear memristive dynamics are included within the computational graph used in the gradient calculation process, enabling more precise fine-tuning of network parameters.

The action potentials generated by memristive neurons are a result of nonlinear ion-driven dynamics, which are functionally equivalent to a chain of nonlinear operations in a computational graph. As each spike is a result of naturally arising physics-driven phenomena rather than discrete switching in digitally-composed SNNs, all functions are fully differentiable and the gradient is well-defined. Therefore, there is no need to resort to surrogate gradient approaches that have become ubiquitous when training SNNs. The voltage action potential is scaled by memristive synapses, and the resultant current drives downstream layers of memristive neurons.

An overview of our approach is shown in Fig.~\ref{integrate} and a summary of our contributions are provided below:

\begin{itemize}
    \item We present a fully memristive SNN utilizing both memristive neurons and synapses, and is shown to achieve state-of-the-art accuracy for lightweight dense network architectures on real-world datasets. These datasets are more complex than what has been demonstrated in the past on MSNNs while using naturally arising physics-driven phenomena in RRAM; 
    \item We propose \textit{MEMprop}, the application of error backpropagation directly to SPICE circuit models of memristive neurons such that higher-order device dynamics are fully utilized in the learning process. Memristive dynamics are broken down into a series of composable, differentiable functions and used during gradient descent;
    \item SNNs are shown to be trainable without the need for gradient approximations around hard thresholds, such as with surrogate gradient descent, and
    \item The need for ADCs is amortized by relying entirely upon analog current injection at the input, and analog action potentials at the output.
\end{itemize}

In doing so, the rich dynamics of nonlinear devices can be fully leveraged in larger systems, in a way that moves far beyond applying fault injection and variation into the training process. We achieve state-of-the-art accuracy for lightweight dense network architectures on several static and neuromorphic datasets, which pushes the fringe of what previous MSNNs have shown to be capable of.

\section{Background}

There is an incredibly broad span of work that integrates memristors with brain-inspired architectures, from low-level analog action potential emulation \cite{kang2021build}, to discrete spiking dynamics \cite{eshraghian2022memristor}, up to non-spiking CIM processors \cite{cai2019fully}. We focus our background on prior work that use nonlinear dynamics in memristive neurons together with memristive synapses, with associated demonstration of synaptic optimization to achieve a data-driven outcome.

\subsection{Fully Memristive SNNs}
Full memristive SNNs refer to arrays that utilize nonlinear switching dynamics of memristors to trigger action potentials, and are coupled with memristive weights that are used as neural network parameters. 
The 8$\times$8 crossbar array presented in \cite{wang2018fully} successfully integrates a fully memristive SNN, including memristive synapses and neurons. The synaptic array is trained using unsupervised STDP to classify four letters in a 24-pixel grid. While the task achieved is considerably simple, the fully memristive experimental demonstration sets the stage to build upon new training methods.

The work in \cite{kiani2021fully} uses half-wave rectification interposed between crossbar arrays to process the ReLU activation in the analog domain. While not fully memristive, nor a `spiking' network, it demonstrates a pivotal example of successively passing analog activations between RRAM crossbars without intermediate data conversion, in a manner akin to how our analog action potentials are transmitted between layers. The training process relies on gradient-based optimization where device non-idealities are injected during the forward pass. In doing so, a test set accuracy of 93.63$\%$ is obtained on the MNIST dataset.

In Refs.~\cite{wang2018handwritten} and \cite{wijesinghe2018all}, convolutional SNNs with memristors are used, where both used pretrained on non-spiking networks that are mapped or converted into the spiking domain. Both networks obtained competitive accuracy on the MNIST dataset, though did not demonstrate performance on more complex, real-world data. This may potentially be due to the large differences between the networks that were trained and the MSNN that was implemented. 
In Ref.~\cite{duan2020spiking}, a dense MSNN is adopted using a similar approach to what is used here, and as such, has minimal hardware requirements at run-time. The training process abstracts away the switching dynamics of the memristive neuron into a firing rate, and may be the reason why a relatively low accuracy of 83.2\% was achieved on the MNIST dataset. A more detailed comparison is illustrated in Section V, subsection B.

Almost all of these works offer compelling demonstrations using in-house fabricated arrays, either with standalone crossbars or back-end-of-the-line (BEOL) integrated arrays with foundry-made chips. In contrast, we have aimed to make our work as device-agnostic as presently possible by using commercially available Knowm memristors and its corresponding model \cite{molter2016generalized}. 

Although Knowm memristors are not known for their reliability, their metastable switching dynamics are accounted for within the gradient calculation step, as it forms part of the computational graph. In contrast, Kiani~\textit{et al.} use the memristors in the forward-pass only, as their devices are not intended to be reprogrammed during inference, and thus do not require switching to generate spiking dynamics. Gradients can therefore be deterministically calculated partially off-chip \cite{kiani2021fully}, whereas our approach harnesses memristive dynamics in the forward-pass computation in our network. As such, our MSNN approach can leverage the benefits of spike-based processing, such as sparse processing and lower data collision rates.

\subsection{Memristive Learning Frameworks}
To ease and emulate the training process of memristive networks, a variety of valuable frameworks have been developed that each address various niches. These include MemTorch \cite{lammie2022memtorch}, NeuroSim \cite{chen2018neurosim}, and the IBM Analog Hardware Acceleration Kit \cite{rasch2021flexible}, which implement non-spiking networks that adopt mixed-signal bit-line charge/current accumulation/summation processing. In these simulators, memristive dynamics are accounted for during weight updates, and are otherwise fixed during inference. To complement these tools, NeuroPack \cite{huang2022neuropack} specifically targets the simulation of spiking networks, where memristive dynamics are also factored in during the weight update process, and fixed during inference. Spiking dynamics are triggered by pulse-based input voltages.

The closest relation to our proposed work comes from Demirag \textit{et al.} who offer a software implementation of a real-time recurrent learning variant \cite{williams1989learning, bellec2020solution} using phase change device models to train a MSNN \cite{demirag2021online}.

Much like these simulators, our approach \textit{MEMprop} integrates SPICE-level memristor models into the training process. But distinct from these simulators and training methods, we utilize arrays of memristive integrate-and-fire (MIF) neuristors that trigger self-induced action potentials during the forward-pass computation. To emit spikes, the state of a memristor must evolve and ultimately switch to provide sudden discharge pathways which triggers spikes. Therefore, the dynamical characteristics of a memristor will alter the gradient itself, whereas all prior approaches account for dynamics in the weight update step when reprogramming memristive synapses.

Our approach applies the backpropagation algorithm directly to SPICE-level neuristor models that emit analog action potentials as a result of charge-accumulation and metastable threshold switching. That is to say, the memristive dynamics in SPICE models are not fixed during the forward-pass, but rather, they are dynamically extracted during the forward-pass to fire neuristor-induced spikes. The BPTT algorithm is applied \textit{directly} to SPICE models, thus integrating memristive neurons as part of the gradient calculation step, rather than isolating their dynamics to synaptic weight updates as in prior implementations. This offers a dynamical and totally new way to train fully memristive networks that uses state-based dynamics as part of the gradient calculation step. 

In terms of hardware implementation, the conventional use of RRAM in circuits often requires significant overhead to convert analog currents into digital voltages and consumes large amount of power \cite{cai2019fully}. In many instances, the power/area demands of the ADCs and and Digital-to-Analog Converters (DAC) far exceeds the overhead brought on by RRAM, offsetting the advantages of memristors. In contrast, our spike-based approach eliminates the need for ADCs and DACs such that the cost of peripheral circuits are substantially reduced.

\subsection{Error Backpropagation Through SPICE models}

Our proposed approach enables us to scale up the complexity of learnable tasks in fully MSNNs by directly applying gradient descent to the nonlinear state evolution of memristive neurons and synapses. Both neurons and synapses in biological neural networks are modeled using memristors. The MIF neuron model is designed to achieve distinct depolarization, hyperpolarization, and repolarization voltage phases with a minimal set of circuit elements. Memristive synapses act as interconnects between layers of neurons.  

To train the fully memristive network, we propose MEMprop, an application of error backpropagation to large-scale networks derived from SPICE circuit models. This enables dynamical, time-varying memristive neurons to learn and thus achieve much higher accuracy on data-driven tasks than has been previously reported with MSNNs. By relying on the analog spiking characteristics that naturally occur in the MIF neuron model, the non-differentiability of spike-based activations are completely avoided. This means MEMprop does not rely on surrogate gradient techniques that are commonly used to train SNNs, which calculates biased gradient estimators to circumvent the dead neuron problem \cite{neftci2019surrogate}. To promote broad accessibility of our methods, we use SPICE models of commercially available, low-cost memristors and demonstrate the efficacy of MEMprop in a supervised deep learning framework.


\section{Methods}
\subsection{Memristive Integrate-and-Fire Model}


\begin{table*}[!ht]
\centering
\caption{MIF model differential equations vs numerical integration.}
\label{miftable}
\begin{threeparttable}
\begin{tabular}{ccc} \toprule \toprule
\textbf{Variable} & \textbf{Continuous Time Derivative} & \textbf{Discrete Time Solution} \\ \midrule
$v$ & 
$\frac{{\mathrm{d}}v}{\mathrm{d}t}=\frac{I - G_1(v-E_{\rm rest}) - G_2(v-E_{\rm reset})}{C} $ &
$v^{t+1} = \frac{I_{t}-G_1^{t}(v^{t}-E_{\rm rest})-G_2^{t}(v^{t}-E_{\rm reset})}{C} + v^{t}$ \\ \midrule
$G_1$ &
$G_1 = \frac{x_1}{R_{\rm on1}} + \frac{1-x_1}{R_{\rm off1}}$ &
$G_1^{t+1} = \frac{x_1^{t+1}}{R_{\rm on1}} + \frac{1-x_1^{t+1}}{R_{\rm off1}}$ \\ \midrule
$G_2$ & 
$G_2 = \frac{x_1}{R_{\rm on2}} + \frac{1-x_2}{R_{\rm off2}}$ & 
$G_2^{t+1} = \frac{x_2^{t+1}}{R_{\rm on2}} + \frac{1-x_2^{t+1}}{R_{\rm off2}}$ \\ \midrule
$x_1$ & 
$\frac{{\mathrm{d}x_1}}{\mathrm{d}t}= \frac{1}{\tau_1}(\frac{1-x_1}{1+\exp(\frac{v_{\rm on1}-(v-E_{\rm rest})}{V_{\rm th} \cdot k_{\rm v}})}  -   \frac{x_1}{1+\exp(\frac{(v-E_{\rm rest})-v_{\rm off1}}{V_{\rm th} \cdot k_{\rm v}})})$ &
$x_1^{t+1} = \frac{1}{\tau_1} \Big(\frac{1-x_1^{t}}{1+\exp(\frac{v_{\rm on1}-(v^{t}-E_{\rm rest})} {V_{\rm th} \cdot k_{\rm v}})}  - 
\frac{x_1^{t}}{1+\exp(\frac{(v^{t}-E_{\rm rest})-v_{\rm off1}}{V_{\rm th} \cdot k_{\rm v}})} \Big)+ x_1^{t} $ \\ \midrule
$x_2$ &
$\frac{{\mathrm{d}x_2}}{\mathrm{d}t}= \frac{1}{\tau_2}(\frac{1-x_2}{1+\exp(\frac{v_{\rm on2}-(v-E_{\rm reset})}{V_{\rm th} \cdot k_{\rm v}})}  -   \frac{x_2}{1+\exp(\frac{(v-E_{\rm reset})-v_{\rm off2}}{V_{\rm th} \cdot k_{\rm v}})}) $ &
$x_2^{t+1} = \frac{1}{\tau_2} \Big(\frac{1-x_2^{t}}{1+\exp(\frac{v_{\rm on2}-(v^{t}-E_{\rm reset})} {V_{\rm th} \cdot k_{\rm v}})}  - 
\frac{x_2^{t}}{1+\exp(\frac{(v^{t}-E_{\rm reset})-v_{\rm off2}}{V_{\rm th} \cdot k_{\rm v}})} \Big)+ x_2^{t} $ \\
\bottomrule \bottomrule

\end{tabular}
 \end{threeparttable}
\end{table*}

\begin{figure}[!htbp]
\centering
\subfloat[]
{
	\includegraphics[scale=.3]{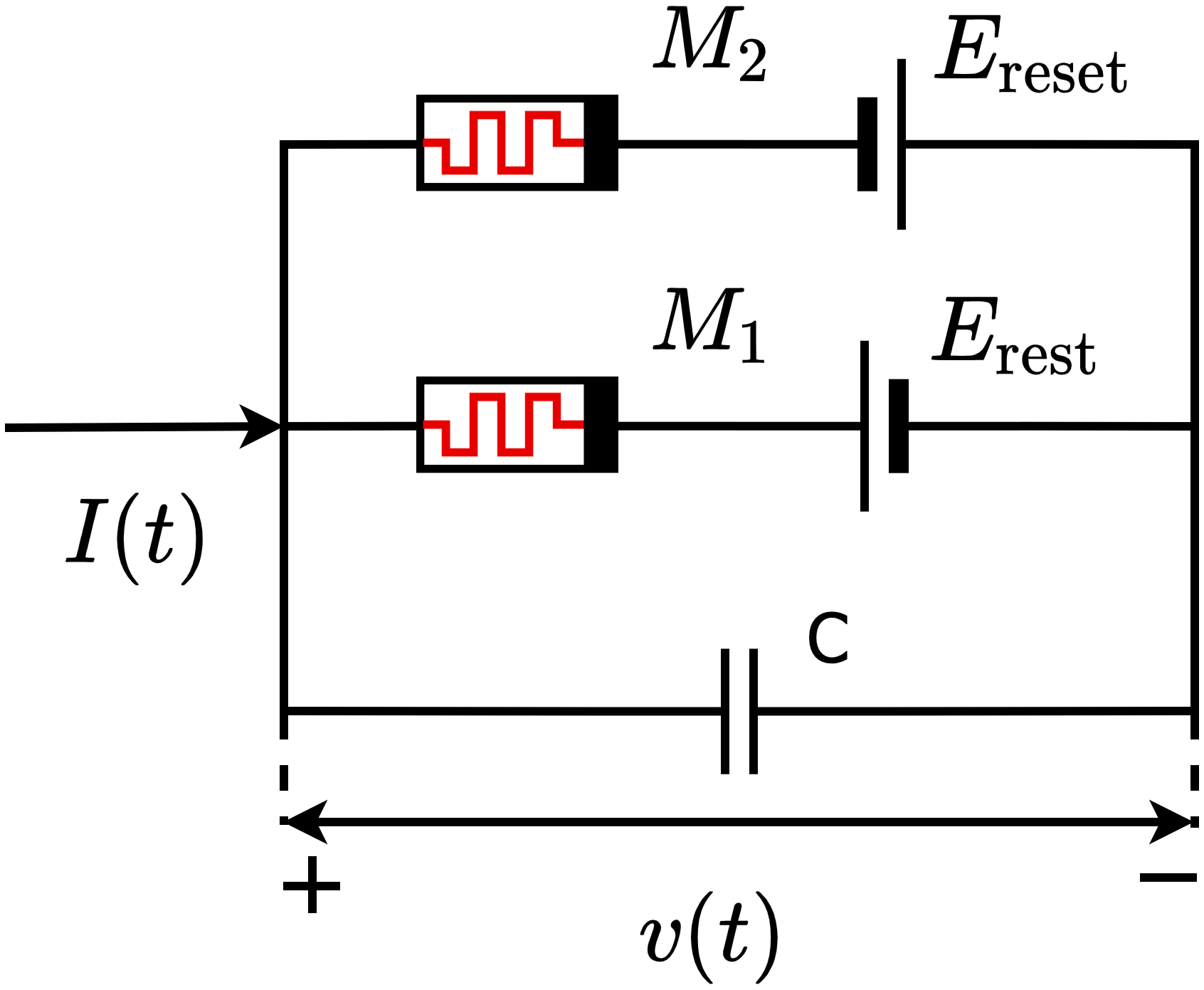}
	\label{fig1a}
}
\subfloat[]
{
	\includegraphics[scale=.08]{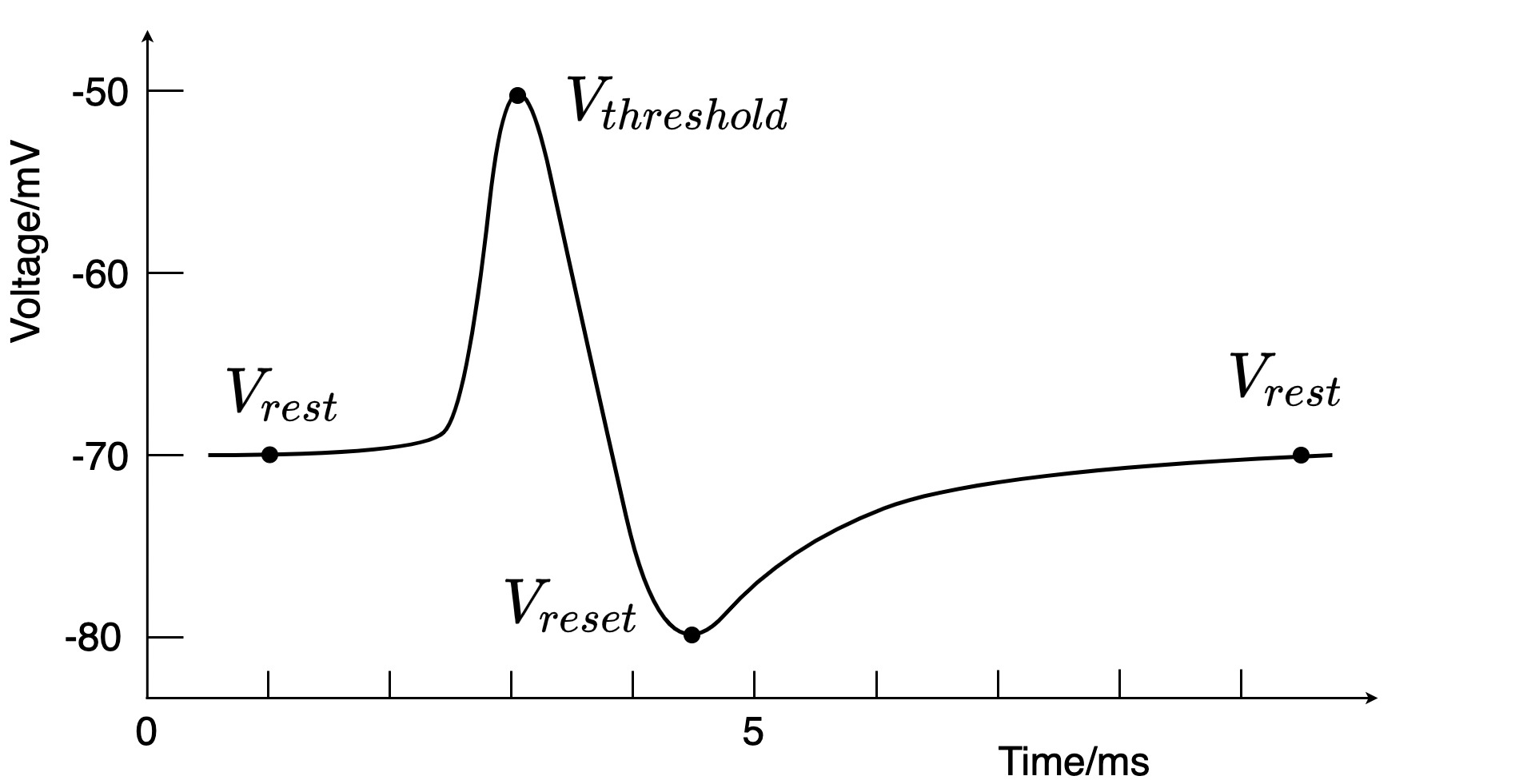}
	\label{fig1b}
}
\caption{(a) A memristive integrate-and-fire (MIF) neuron consists of two memristors $M_1$ and $M_2$, connected to DC voltage sources $E_{\rm rest}$ and $E_{\rm reset}$, in parallel with a capacitor $C$. The MIF neuron is provably minimal in generating a membrane potential traversing from a rest voltage level to a threshold voltage level, then to a reset voltage level, and then back to the rest potential when a current pulse is applied. (b) An action potential waveform that MIF model resembles showing rest, threshold and reset potentials.}
\label{mif2circuit}
\end{figure}

The neuron model adopted in our MSNN is the MIF neuron model depicted in Fig.~\ref{mif2circuit}(a) \cite{kang2021build}. Qualitatively, given a positive current injection $I(t)$, the membrane potential $v(t)$ will charge up from the neuron's resting potential $E_{\rm rest}$. Once a sufficiently large electric field builds up across the memristor $M_2$, it switches on, effectively shorting the output to $E_{\rm reset}$ which charges back up to equilibrium, $E_{\rm rest}$. The evolution of the membrane potential is shown in Fig.~\ref{mif2circuit}(b), and our prior work in Ref.~\cite{kang2021build} provides an experimental demonstration using a pair of Knowm memristors.

The memristor model used is based on the generalized metastable switch (MSS) model \cite{molter2016generalized}, which has been used to accurately describe a large range of possible devices. In this model, a MSS is an idealized element that switches with a given probability between two states as a function of its voltage and temperature. A memristor is modeled by a collection of MSSs, which determines the state-time dynamics that lead to non-volatile characteristics. $x$ characterizes the internal state as a variable normalized between 0 and 1, as determined by the switching states of all MSSs. 

Formally, the governing dynamics of the MIF neuron are characterized by the system of differential equations in Table~\ref{miftable}. The membrane potential $v$ is dependent on $G_1$ and $G_2$, the device conductances of $M1$ and $M2$, the MIF capacitance $C$, and $E_{\rm rest}$ and $E_{\rm reset}$, which are voltage biases in the MIF circuit.

$G_1$ and $G_2$ are dependent on $x_1$ and $x_2$, a pair of variables governing the internal state of each device, where $R_{\rm on}$ and $R_{\rm off}$ are the on/off resistances. 

The rates of change of $x_1$ and $x_2$ are dependent on the state evolution time constants $\tau_1$ and $\tau_2$, the thermal voltage $V_{\rm th}$ at room temperature, and the volatility constant $k_{\rm v} \in [0, 1]$ in the range of 0 to 1. As $k \rightarrow 0$, the more retentive the device is. 


\subsection{Neural Network Layout}
When accounting for the hardware implementation of a fully memristive neural network, the voltage response of the MIF neuron must drive the memristive synapses, and is correspondingly weighted by the synaptic conductances. This is can be fully integrated in a crossbar according to Eq.~(\ref{rramohm}) below:

\begin{equation}\label{rramohm}
\mathbf{I} = \mathbf{G} \times \mathbf{V}
\end{equation}

\begin{align*}
\begin{bmatrix}
  I_{1}  \\
  I_{2}  \\
  \vdots \\
  I_{n}
\end{bmatrix} &= \left[\begin{array}{cccc}
  w_{00} & w_{01} & \cdots & w_{0m} \\
  w_{10} & w_{11} & \cdots & w_{1m} \\
  \vdots & \vdots & \ddots & \vdots \\
  w_{n0} & w_{n1} & \cdots & w_{nm} \\ 
\end{array}\right]
\begin{bmatrix}
  v_{1}  \\
  v_{2}  \\
  \vdots \\
  v_{m}
\end{bmatrix} \\
\end{align*}

\begin{equation}\label{ivg}
I_{n} = \sum_{i=1}^m v_{i} \times w_{ni}
\end{equation}

\noindent where $\mathbf{I}$ is the output current vector, $\mathbf{V}$ is the input voltage vector, and $\mathbf{G}$ is the conductance matrix of the crossbar. The output current vector is generated via bit-line current summation as shown in Eq.~(\ref{ivg}), and directly drives the input of the next MIF neuron layer. Resistive loading may attenuate the output current in subsequent stages, but this can be accounted for using a scaling factor, or otherwise buffered \cite{wang2020high, wang2022low, kang2021build}.

\begin{figure}[!htbp]
\centering
\includegraphics[scale=0.45]{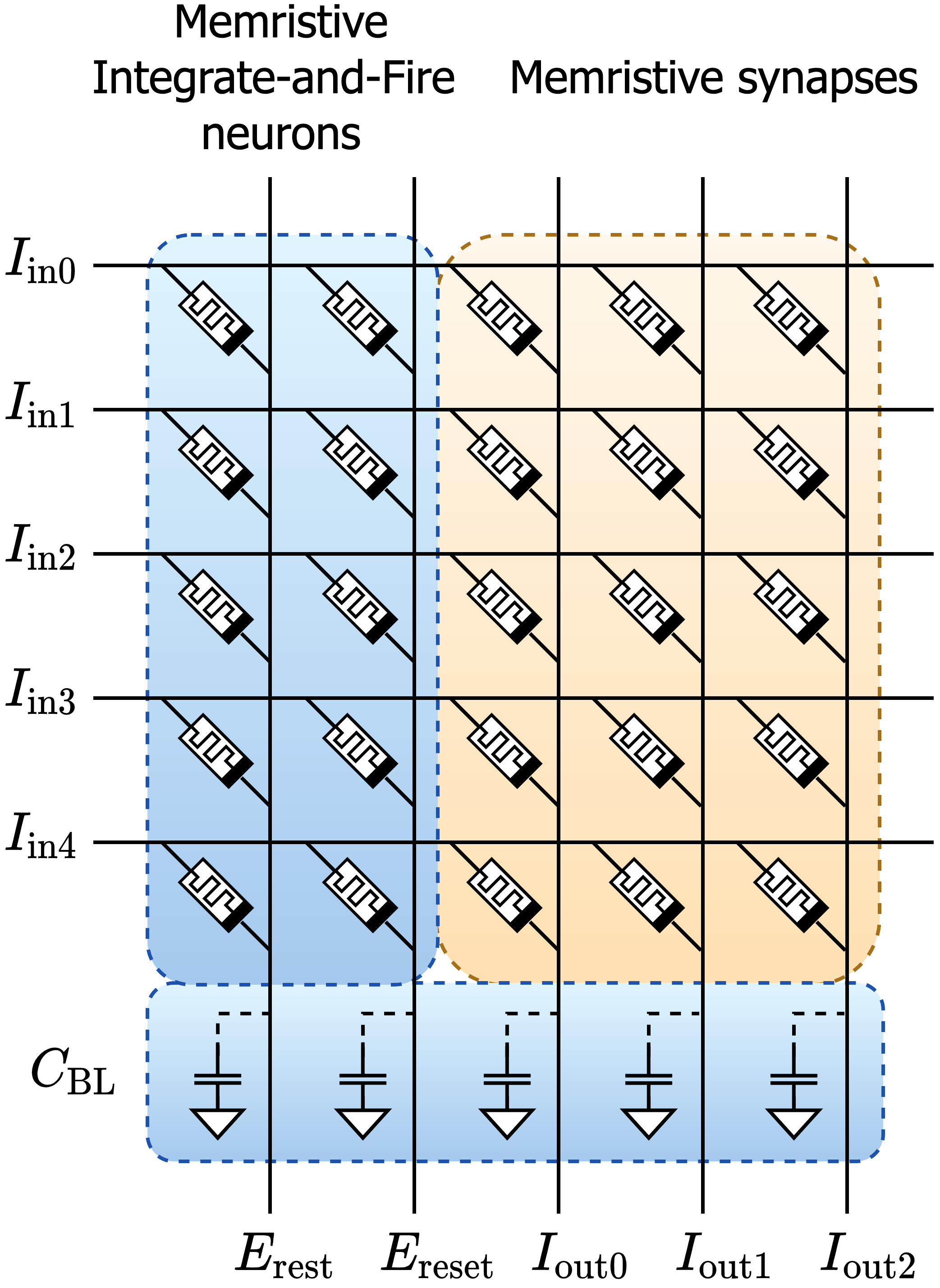}
\caption{Schematic of a fully memristive neural network. $C_{\rm BL}$ represents the parasitic bit-line capacitance.}
\label{crossbar}
\end{figure}

Fig.~\ref{crossbar} shows a small-scale schematic of a fully memristive neural network with five MIF neurons and a $5 \times 3$ memristive crossbar, which receives five fan-in input currents $I_{\rm in0}$ - $I_{\rm in4}$ and generates three fan-out output currents. $E_{\rm rest}$ and $E_{\rm reset}$ are the voltage sources in the MIF circuit. $C_{\rm BL}$ is the bit-line parasitic capacitance generated by the metal line, or otherwise by the drain-bulk capacitance of select transistors (not shown), and is used as the membrane capacitance in the MIF circuit.

\subsection{Forward-mode Solution of MIF Model}
In order to train a network of MIF neurons and synapses using gradient descent, the differential equations representing the MIF circuit dynamics (middle column, Table~\ref{miftable}) are recast into discrete-time form (left column, Table~\ref{miftable}). In doing so, the memristive dynamics can be captured in a computational graph that evolves over time, much like a recurrent neural network (RNN). 

In practice, SPICE simulators use a variety of differential equation solvers, such as the backward Euler method, and the 4th-order Runge-Kutta method (RK4). For compatibility with the BPTT algorithm, we solve the differential equations using the forwrd Euler method, which provides an explicit representation of the next time step using present-time dynamics. The rich dynamics of the MIF neuronal network are now accounted for in the MSNN, unrolled in time such that gradient descent can be used to optimize the memristive synapses as a function of the MIF evolution.

Many neural coding studies represent spike trains as a summation of time-shifted Dirac delta pulses $\sum_n{\delta(t-t_j^n)}$. As such a model of spikes are an idealization, we use the spike train to modulate a time-continuous, alpha-shaped input current $I$ modeled by the equation in Table~\ref{alphatable} to be compatible with real, physical systems. In this series of equations, $a$ is an internal state variable, $\tau_{\rm syn}$ is a time constant that determines the shape of the alpha current, $W_i$ is the synaptic weight between a pre-synaptic neuron $i$ and its associated post-synaptic neuron. It is commonly regarded that such alpha-shaped waveforms correspond to the response from biological neurons in the sensory periphery that respond with graded potentials \cite{eshraghian2018formulation}. 


\begin{table*}[!ht]\centering \caption{Alpha current differential equations vs numerical integration.}\label{alphatable}
\begin{threeparttable}
\begin{tabular}{ccc} \toprule \toprule
\textbf{Variable} & \textbf{Continuous Time Derivative} & \textbf{Discrete Time Solution} \\ \midrule
$I$ & 
$\tau_{\rm syn} \frac{{\mathrm{d}I}}{\mathrm{d}t} = a-I $ &
$I^{t+1} = \frac{a^{t}-I^{t}}{\tau_{\rm syn}} + I^{t}$ \\ \midrule
$a$ &
$\tau_{\rm syn} \frac{{\mathrm{d}a}}{\mathrm{d}t} = -a + W_i \cdot \sum_n{\delta(t-t_i^n)}$ &
$a^{t+1} = -\frac{a^{t}+W_i \cdot \sum_n{\delta(t-t_i^n)}}{\tau_{\rm syn}} + a^{t}$ \\
\bottomrule \bottomrule
\end{tabular}\label{equations}
 \end{threeparttable}
\end{table*}

\subsection{MIF Single Neuron Simulation}
\begin{figure}[!htbp]
\centering
\subfloat[]
{
	\includegraphics[width=3in]{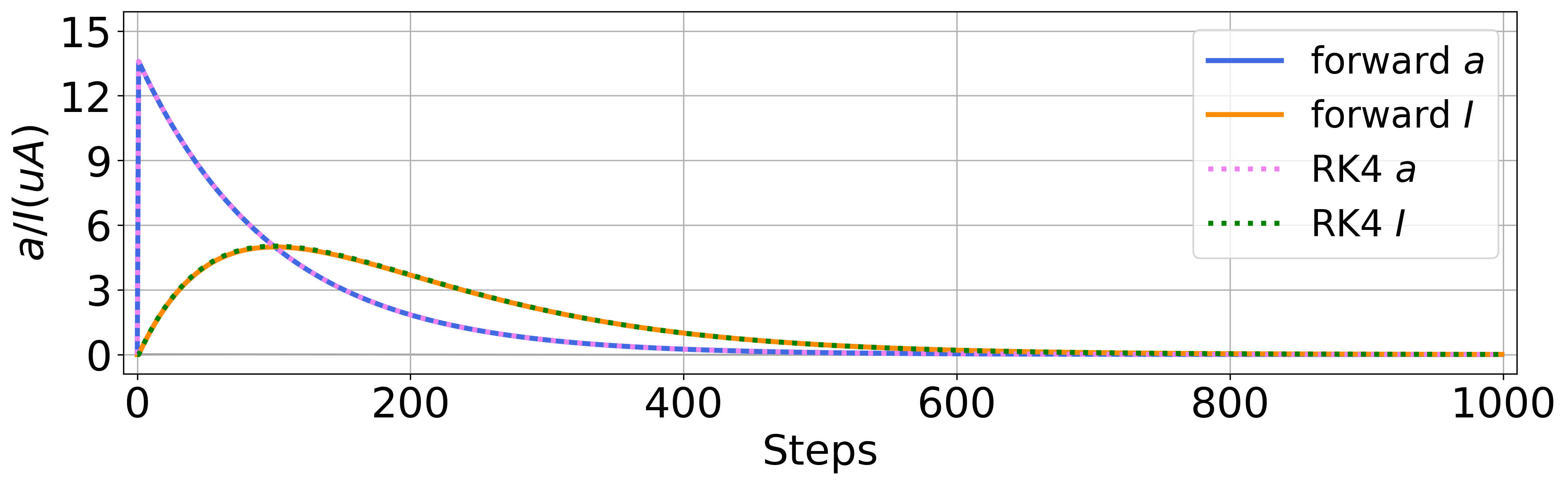}
	\label{mnist}
}\\
\subfloat[]
{
	\includegraphics[width=3in]{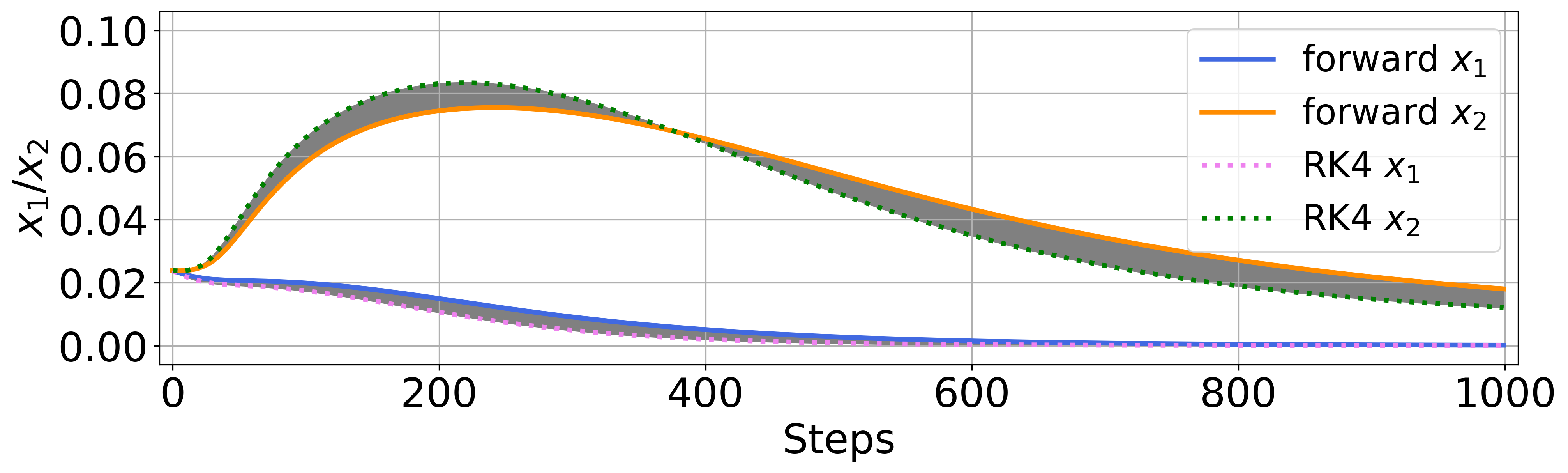}
	\label{fmnist}
}\\
\subfloat[]
{
	\includegraphics[width=3in]{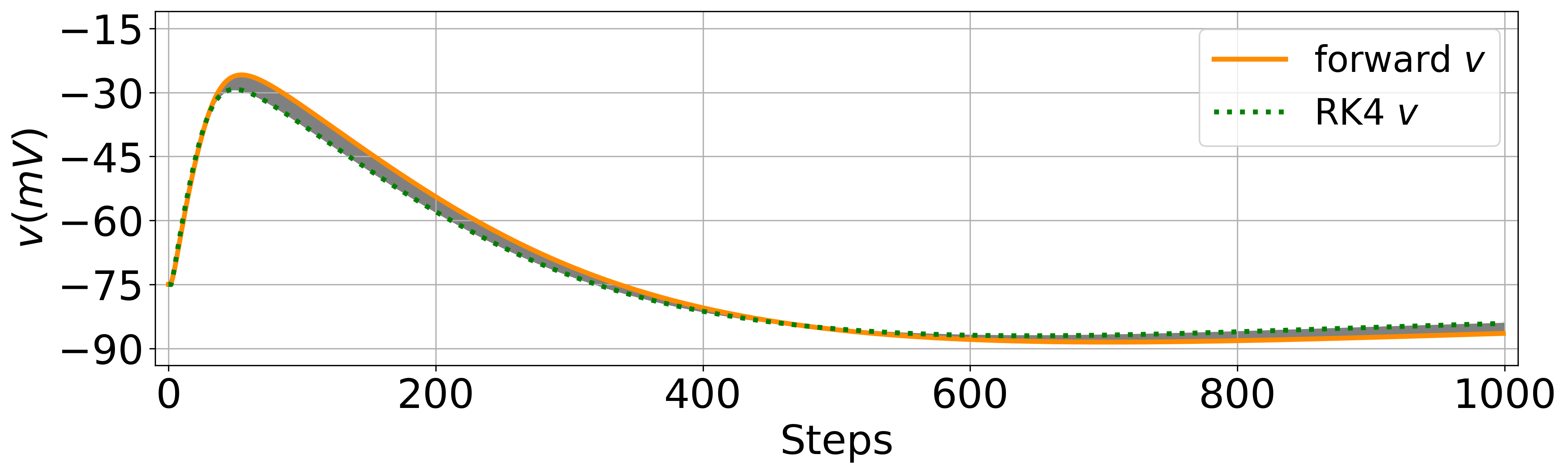}
	\label{fmnist}
}
\caption{Simulation results of MIF neuron solved using forward Euler numerical integration. The results matches the quantitative dynamics of solved by a SPICE simulator. Difference between two methods is marked by grey area. (a) Alpha synaptic current dynamics. (b) Internal states $x_1$, $x_2$. (c) Voltage response $v$.}
\label{mif2Euler}
\end{figure}

To verify the accuracy of the forward Euler method, we conduct a single neuron simulation using Python shown in Fig.~\ref{mif2Euler} across 1,000 time steps, which provides ample time for the membrane potential $v$ to traverse from spiking $V_{\rm th}$, to the reset potential $E_{\rm reset}$, back to the resting potential $E_{\rm rest}$. Each time step is of duration 0.01 ms over a total duration of 10 ms. 

The parameters used in the simulation are listed in Table~\ref{tablemsnnparameter}. The resting potential, charge integration, thresholding, and reset dynamics are closely reproduced in a SPICE simulator that uses the RK4 solver, which verifies the solution generated by adopting the forward Euler method.
Deep learning is known to be tolerant to fault injections \cite{lammie2021memristive, pentecost2021nvmexplorer}, and so the tradeoff between the numerical integration accuracy and the training complexity should be considered. We adopted Forward Euler method which is able to balance the two. Technically, our training method is expected to generalize to other numerical integration methods as well, but with added computational complexity.

\begin{table}[!htbp]
\caption{MIF circuit parameters}
\label{tablemsnnparameter}
\begin{center}
\begin{tabular}{|c|c|c|c|}
\hline
\textbf{Parameter} & \textbf{Value} & \textbf{Parameter} & \textbf{Value} \\ \hline 
$v_{\rm off_1}$, $v_{\rm off_2}$ & 5 mV & $v_{\rm on_1}$, $v_{\rm on_2}$ & 110 mV \\
$R_{\rm off_1}$, $R_{\rm off_2}$ & 0.1 M$\Omega$ & $R_{\rm on_1}$, $R_{\rm on_2}$ & 1 k$\Omega$ \\
$\tau_1$, $\tau_2$ & 1 ms & $\tau_{\rm syn}$ & 0.64 ms \\
$E_{\rm rest}$ & 0 mV & $E_{\rm reset}$ & 50 mV \\
$V_{\rm th}$ & 25 mV & $k_{\rm v}$ & 0.6 \\
$C$ & 100 pF & & \\
\hline
\end{tabular}
\end{center}
\footnotesize{}
\end{table}

\subsection{BPTT in MSNNs}
The discrete-time solution in Table~\ref{miftable} is illustrated as a directed, acyclic graph in Fig.~\ref{integrate}, where time flows from left to right. The MIF circuit parameters are color-coded to show how each electrical characteristic impacts the others at the next time step.

To train a network, a loss function is calculated using the membrane potential $v$ of the output layer at each step. Note that the adjoint method in \cite{chen2018neural} is not adopted here, as an intermediate state is required to calculate loss and guide training for each time step. We are concerned with the spiking output at all time steps, and not just the final state of the system which makes BPTT a more optimal choice. In our example network provided in Fig.~\ref{integrate} with 10 output neurons, the predicted MIF neuron is expected to spike most frequently by aiming to increase the membrane potential across time steps, while the incorrect target should be suppressed. As the membrane dynamics are continuous, a fully analog MSNN can be trained without surrogate gradients, as has become ubiquitous in deep SNNs trained via error backpropagation \cite{neftci2019surrogate, eshraghian2021training}.

The BPTT algorithm iteratively applies the chain rule from the output back to the leaf nodes, $w$, to determine the update direction to optimize the network. Prior demonstrations treat $w$ as the device conductance. In our case, we also do this, but additionally generate spiking behaviors using naturally occuring MIF dynamics rather than applying a hard threshold to the membrane potential. Not only is this a more biorealistic representation, but it can be fully integrated using RRAM crossbars.


\section{Experimental Results}

\subsection{Network Architecture}
To validate the use of MIF neurons in a deep learning framework, we used a lightweight 3-layer dense SNN with 100 hidden MIF neurons (bottom-left of Fig.~\ref{integrate}). Each MSNN is simulated for a duration of 10~ms over a span of 1,000 discrete time steps.

The input layer consists of a number of input features (784 for the static datasets; 2,048 for a downsampled neuromorphic dataset). An alpha current generator with an amplitude weighted by the input pixel intensity (Table~\ref{alphatable}), weighted by memristive synapses, inducing 100 MIF neurons to fire into another set of memristive synapses, terminated by the output layer of MIF neurons. To account for circuit loading effects, the current injected to each MIF layer is attenuated by a scaling factor defined in the hyperparameters. 


\subsection{Datasets}
Three datasets of increasing difficulty are used to assess the MSNN: MNIST, FashionMNIST, and the DVS128 Gesture datasets. Despite being considered a `solved' problem for quite some time now, it was often the case that the MNIST dataset was the most challenging problem that could be solved by previously reported MSNNs. In most cases, a subset or simplified alternative would be used to demonstrate pattern recognition. Here, we demonstrate for the first time that RRAM arrays that rely on internal dynamics are still capable of processing more challenging problems and real-world datasets, namely, FashionMNIST and neuromorphic data.

The MNIST~\cite{lecun1998gradient} and the slightly more challenging FashionMNIST datasets~\cite{xiao2017fashion} are used to test the performance of the MSNN on temporally static data. Both datasets consist of 60,000 28$\times$28 greyscale images in the training set, and 10,000 images in the test set, with 10 output classes of handwritten digits (MNIST) and clothing items/accessories (FashionMNIST/FMNIST). During the training process, the alpha current inject is applied at every 100 steps to promote sparse network activity.

For a more challenging test case, the DVS128 Gesture dataset~\cite{amir2017low} is used a neuromorphic baseline to test the MSNN's performance on event-driven data filmed with an event-based camera~\cite{patrick2008128x}. Each sample only processes sufficient changes in luminance, and consists of 11 different output classes of hand gestures, such as clapping, arm rotation, and air guitar. Each sample is downsampled to a resolution of $32 \times 32 \times 2$, where the channel depth of 2 accounts for on and off spikes (positive and negative luminance changes). The training data is integrated to fit within 100 discrete time steps, and the testing data is integrated within 360 time steps, to account for GPU memory constraints.

\subsection{Training Process}
The forward Euler solution of each MIF neuron was defined in PyTorch v1.10.1 in Python 3.8, where the autodifferentiation framework was used to keep track of gradients of all forward-mode computations on-the-fly.

The loss used is the negative log-likelihood of the membrane potential $v(t)$ (Fig.~\ref{integrate}) of the output layer. The network objective is to increase voltage across the MIF neuron associated to the correct class. The total loss across time-steps is summed prior to backpropagating the error through the SPICE memristor model using the Adam optimizer \cite{kingma2014adam}. The MSNNs are each trained for 50 epochs.



\subsection{Results}
\begin{figure}[!htbp]
\centering
\includegraphics[width=3.3in]{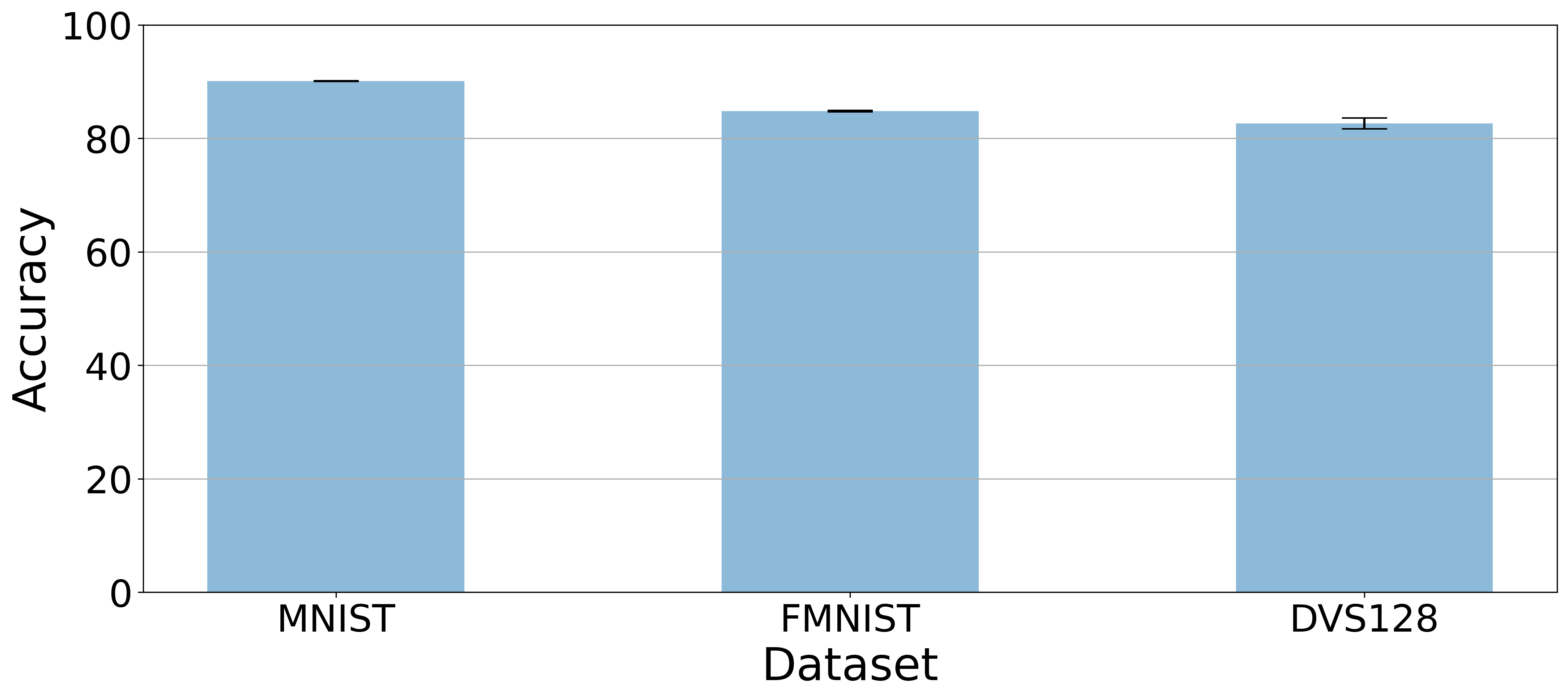}
\caption{Test set accuracies with error bars over 5 trials.}
\label{accuracy}
\end{figure}

The final test set accuracy is measured across 50 epochs over 5 trials for each dataset with early stopping applied, and the average result is shown in Fig.~\ref{accuracy} with error bars.

The average accuracy on the MNIST dataset reaches 93.08\%, 
which is very high among fully MSNN considering almost no required peripheral circuitry for hardware implementation and far exceeds other MSNN baselines that have been previously reported. Although it remains somewhat below non-memristive baselines which can obtain greater than 97\% on the MNIST dataset using a similar architecture, such models are typically computed using fixed- and full-precision arithmetic rather than depending on naturally arising memristive dynamics.

Where our approach tends to shine is on real-world data that goes beyond the simplicity of handwritten digit recognition. Our fully MSNN performance holds for the FashionMNIST dataset, where we obtained an average of 84.77\%, and also for the DVS128 Gesture dataset, with 82.63\%. This provides the first successful demonstration of training a MSNN on a neuromorphic dataset by relying on naturally occurring device dynamics for spike emission.

\section{Discussion and Conclusion}
\subsection{Area, Power, and Latency}
With our fully MSNN approach, one of the main advantages is that we avoid the need for ADCs, and subsequently, we do not need to multiplex or serialize data. To quantify the potential improvement in terms of power, area, and latency, we assume a dense 3-layer architecture of 784-100-10 neurons. Such a network requires 110 MIF neurons and 79.4k memristive synapses. For each synaptic weight, a pair of devices are required to implement current subtraction to represent both positive and negative weights. Given the array structure in Fig.~\ref{crossbar}, a total of 0.16M RRAM cells are required, which needs 10 RRAM tiles of 128$\times$128 in size.


To provide an area estimation at a 65-nm technology node, the size of each RRAM tile with 1T1R cells is 2.77×10$^{-3}$mm$^{2}$, and a complete array will occupy 2.77×10$^{-2}$mm$^{2}$ \cite{wang2019deep}. If a bit-line current summation approach was adopted for a dense ANN with 8-bit networks, a current-mode SAR ADC each occupies 3×10$^{-3}$mm$^{2}$, where 4 ADCs are shared across the column wires for each crossbar. In such a case, the improvement of our approach without converting and serializing activations by adopting a spike-based approach is a factor of 5.33$\times$.


Estimating the power consumption of asynchronous, spike-based workloads requires profiling the spiking activity in the network for a given dataset. Using the MNIST dataset, we found that an average of 2\% of neurons are spiking at any given time in the network. This may be modulated to promote varying degrees of excitation in the network, for example, by applying a spike-dependent regularization term to the objective function. The average resistance in our network is close to the midpoint between the on/off resistances: $R_{\rm off}$ = 100 k$\Omega$, $R_{\rm on}$ = 1000 $\Omega$, average $R_{\rm ave}$ = 50.5 k$\Omega$. The voltage across each device when emitting a spike follows a sharp peak of $v_{\rm on}$ = 110 mV, a refractory period where the potential drops down to $v_{\rm off}$ = 5 mV, and the average value assuming the spike is linearized for the window of interest is $v_{\rm ave}$ = 57.5 mV can give an estimate of the average power 

\begin{equation}
    P_{\rm ave} = \frac{v_{\rm ave}^2}{R_{\rm ave}} \times N_{\rm cells} \times 2\%,
\end{equation}

\noindent which is then numerically evaluated for a single 128$\times$128 tile to be 21.45$\mu$W, and 0.21mW when accounting for all 10 tiles. When accounting for data conversion overhead where each 8-bit ADC consumes 2×10$^{-4}$W, the total power consumption increases to 8.21~mW. Our approach can offer an improvement of 38.30$\times$.



We estimate latency by driving ADCs at an operating frequency of 40MHz. With 4 ADCs converting 64 column currents (assuming 2 bit-lines per activation), 32 data-lines are grouped to share a driving DAC which generates an 8-bit serialized input operating across 8 distinct cycles, evaluating to input data latency of 6.4$\mu$s. At the output, 32 bit-lines each share an ADC where 16 column currents must be converted, increasing the latency by 0.4$\mu$s. Further assuming that data conversion dominates latency, this means that it takes 6.8$\mu$s per vector matrix multiplication (VMM). A 1,000 time step simulation, as used in our network, will extend this to a latency of 6.8~ms. Additionally, each input is fed as an alpha current signal with a measurable latency around 0.64~ms, giving the total latency for the conventional approach of 7.44~ms, as opposed to MEMprop which only requires the additional switching time of the memristors (on the order of 100s of nanoseconds added to the alpha current latency). In our approach, currents and voltages are generated without the need for conversion or serialization, and so there is no added ADC latency. The alpha current injection and response already account for RC delay through the memory array.

A summary of the above mentioned comparison is shown in Table~\ref{tableimprovement}.


\begin{table}[!htbp]
\caption{Power, area, and latency improvement in our MSNN}
\label{tableimprovement}
\begin{center}
\begin{tabular}{l|l|l|l}
\toprule \toprule
\textbf{Aspect} & \textbf{Our method} & \textbf{Mixed-Signal} & \textbf{Improvement} \\ \hline 
Area & 2.77 mm$^{2}$ & 14.77 mm$^{2}$  & ~5.33× \\
Power & 0.21 mW & ~8.21 mW & 38.30×\\
Latency & 0.64 ms & ~7.44 ms & 11.63x\\
\bottomrule \bottomrule
\end{tabular}
\end{center}
\footnotesize{}
\end{table}

\subsection{Comparison}

A comparative analysis against other memristive networks is provided in Table~\ref{papers_comparison}. Where metrics are not provided in the original sources (e.g., chip area, power consumption, and inference latency), we have made an estimate where sufficient data has been provided to extrapolate these values. The closest MSNNs to our work are briefly described below. 

In Ref.~\cite{wang2018handwritten}, a convolutional neural network with analog neurons and digital spiking at the input layer is adopted. Each analog neuron consists of four transmission gates, seven operational amplifiers, a comparator, and a memristor. While this is substantially more complex than the MIF neuron, the large overhead per neuron is offset by using time-division multiplexing access (TDMA) to treat a single physical neuron as multiple algorithmic neurons.

\newcolumntype{g}{>{\columncolor{Gray}}c}
\begin{table*}[!htbp]
\centering
\caption{Comparison among fully memristive spiking neural networks}\label{papers_comparison}
\begin{threeparttable}
\begin{tabular}{cccccccc} \toprule \toprule
\multirow{2}{*}{\textbf{Method}} & \multirow{2}{*}{\textbf{Architecture}} & \multirow{1}{*}{\textbf{Neuron No.}} & \multirow{2}{*}{\textbf{Refractory}} & \multirow{2}{*}{\textbf{Neuron feature}} & \multirow{2}{*}{\textbf{Training Method}} & \multirow{2}{*}{\textbf{MNIST}} & \multirow{1}{*}{\textbf{Complexity of Additional}} \\
& & \multicolumn{1}{l}{\textbf{of elements}} & & & & & \multirow{1}{*}{\textbf{ Control Logic}} \\ \midrule

\multirow{2}{*}{\cite{wang2018handwritten}} & \multirow{1}{*}{1Mem-C-M-C-} & \multirow{2}{*}{4(5)+33T} & \multirow{2}{*}{\ding{55}} & \multirow{2}{*}{Mixed-signal} & \multirow{1}{*}{LIF rate fit} & \cellcolor[gray]{0.9} & \multirow{2}{*}{high} \\
& M-1024F-10F & & & & \multirow{1}{*}{TDMA} & \multirow{-2}{*}{\cellcolor[gray]{.9}97.1\%} & \\ \midrule
 
\multirow{2}{*}{\cite{wijesinghe2018all}} & \multirow{2}{*}{784-C-M-C-M-10F} & \multirow{2}{*}{5+2T} & \multirow{2}{*}{\ding{55}} & \multirow{2}{*}{Digital} & \multirow{1}{*}{Stochastic switch fit} & \cellcolor[gray]{0.9} & \multirow{2}{*}{high}  \\
& & & & & \multirow{1}{*}{ANN converts SNN} & \multirow{-2}{*}{\cellcolor[gray]{.9}$\sim$96\%} & \\ \midrule
 
\multirow{2}{*}{\cite{duan2020spiking}} & \multirow{2}{*}{784-100F-10F} & \multirow{2}{*}{3} & \multirow{2}{*}{\ding{55}} & \multirow{2}{*}{Analog} & \multirow{1}{*}{LIF shape fit} & \cellcolor[gray]{0.9} & \multirow{2}{*}{low}\\
& & & & & \multirow{1}{*}{surrogate gradient} & \multirow{-2}{*}{\cellcolor[gray]{.9}83.2\%} & \\ \midrule
 
\multirow{2}{*}{Our work} & \multirow{2}{*}{784-100F-10F} & \multirow{2}{*}{3(5)} & \multirow{2}{*}{\ding{51}} & \multirow{2}{*}{Analog} & \multirow{1}{*}{MIF direct train} & \cellcolor[gray]{0.9} & \multirow{2}{*}{low} \\ 
& & & & & \multirow{1}{*}{MEMprop} & \multirow{-2}{*}{\cellcolor[gray]{.9}93.2\%} & \\ \bottomrule \bottomrule
\end{tabular}
\begin{tablenotes}
    \item[1] T: transistor. $^2$ Mem: memristor. $^3$ C: convolutional layer. $^4$ M: max pooling layer. $^5$ F: fully connected layer.
\end{tablenotes}
\end{threeparttable}
\end{table*}

\begin{table}[!ht]
\centering
\caption{Test set accuracies for the MNIST, FashionMNIST, and DVS128 Gesture Datasets.}\label{main_results}
\begin{threeparttable}
\begin{tabular}{cccccc} \toprule \toprule
\multirow{2}{*}{\textbf{Dataset}} & \multirow{2}{*}{\textbf{Batch}} & \multirow{2}{*}{\textbf{LR}} & \multirow{2}{*}{\textbf{Best}} & \multirow{2}{*}{\textbf{Avg. ($n=5$)}} & \multirow{2}{*}{\textbf{$\sigma$}}\\
 & & & & & \\
 \midrule
MNIST & 128 & 1e-4 & \textbf{93.18} & \textbf{93.08} & 0.07  \\ \midrule
FMNIST & 128 & 1e-4 & \textbf{84.79} & \textbf{84.77} & 0.13  \\ \midrule
DVS-128 & 16 & 1e-4 & \textbf{83.21} & \textbf{82.63} & 0.95  \\ \bottomrule \bottomrule
\end{tabular}
\begin{tablenotes}
    \item[1] LR: learning rate $\eta$. $^2$ $\sigma$: standard deviation.
\end{tablenotes}
\end{threeparttable}
\end{table}


Wijesinghe \textit{et al.} also adopt a convolutional SNN using stochastic switching to implement probabilistic firing, which is a natural fit for devices that switch based on random processes \cite{wijesinghe2018all}. A non-spiking ANN is first trained and subsequently converted to an SNN. This approach is known to perform optimally for static datasets on reasonably deep SNNs, but it sets an upper-limit of performance such that the SNN accuracy will not surpass the accuracy of the ANN. Furthermore, ANN to SNN conversion is yet to show success on neuromorphic datasets \cite{eshraghian2021training}. The total area of the design is reported to be 3~mm$^{2}$, and the latency for a single spike is 0.21~$\mu$s. 


In Ref.~\cite{duan2020spiking}, a 3 element neuron is constructed, consisting of a resistor, a memristor, and a capacitor, which is closest in spirit to our own approach here. The neuron is parameterized to fit to a leaky integrate-and-fire neuron model, which relies on hard-thresholded spike generation which is a non-differentiable function. The problem is circumvented by utilizing surrogate gradient descent, and results in a test set accuracy of 83.2\%. This comparatively lower accuracy highlights the challenges of adopting neuristor-like dynamics, and the difficulty of mapping conventional training methods, both supervised (surrogate gradient descent) and unsupervised (STDP), to dynamical RRAM arrays. We expect the lack of a direct correspondence between thresholded leaky integrate-and-fire neurons and memristive neuron spiking is what resulted in the accuracy degradation, though it sets the stage for developing new training methods that account for dynamical, continuous-time switching into the computational graph of the network. In absence of power consumption metrics, we apply the same assumptions as in our MSNN architecture, namely, that 2\% of the network is active at any given time, and estimate the total power consumption is 0.63~W.



No surrogate gradients or associated approximations are required due to the continous-time nature of the spiking behavior in our experiments. The absence of ADCs/DACs in conventional full-precision/fixed-precision bit-line current summation approaches reduces the computational overhead. This is due to the use of \textit{MEMprop}. More details of our experiments are listed in Table~\ref{main_results}.

\subsection{Concluding Remarks}
To the best of our knowledge, this is the first work that has mapped low-level, analog SPICE dynamics directly into an acylic computational graph that is compatible with the backpropagation algorithm. Directly mapping the behavior of dynamical RRAM arrays into an autodifferentiation framework has enabled us to achieve promising performance on real-world datasets on fully memristive SNNs, beyond simple pattern recognition and handwritten digit classification. By harnessing a blend of the intrinsic behaviors of memristors and the sparse network activity of SNNs, the power consumption and latency of our approach surpasses that of all other similar methods without compromising on accuracy.



We propose MEMprop with a lightweight dense neural network to validate this idea of applying backpropagation directly to SPICE circuit model. In the future, a deeper convolutional MSNN can be explored to further improve accuracy and to solve more complex tasks. In such a case, the benefits of MEMprop in terms of area, power, latency should still hold.
Our approach can account for fault injections by incorporating device-to-device variability at the SPICE model level, which are then ported into gradient-based updates. This is important in the commercial memristors available, as learning heterogeneous device dynamics can improve the representational capacity of MSNNs. \textit{MEMprop} should be able to offer a method by which these variations are accounted for during the training process. 

While variation is tolerable, neuristor-based firing relies on device switching which is a limiting factor in RRAM cells that exhibit limited cycle endurance. This motivates the development of devices that have a wide range of behaviors, but are resilient to consistent switching. Nonetheless, we successfully trained our network to be sparsely activated, which reduces the frequency of switching and relaxes this stringent demand on large device endurance.

\ifCLASSOPTIONcaptionsoff
  \newpage
\fi

\bibliographystyle{IEEEtran}
\bibliography{references}

%






\end{document}